\documentclass{ceurart}

\usepackage{listings}
\usepackage{mdframed}
\begin{document}

\copyrightyear{2024}
\copyrightclause{Copyright for this paper by its authors. Use permitted under Creative Commons License Attribution 4.0 International (CC BY 4.0).}

\conference{In: D. Mendez, A. Moreira, J. Horkoff, T. Weyer, M. Daneva, M. Unterkalmsteiner, S. Bühne, J. Hehn, B. Penzenstadler, N. Condori-Fernández, O. Dieste, R. Guizzardi, K. M. Habibullah, A. Perini, A. Susi, S. Abualhaija, C. Arora, D. Dell’Anna, A. Ferrari, S. Ghanavati, F. Dalpiaz, J. Steghöfer, A. Rachmann, J.  Gulden, A. Müller, M. Beck, D. Birkmeier, A. Herrmann, P. Mennig, K. Schneider. Joint Proceedings of REFSQ-2024 Workshops, Doctoral Symposium, Posters \& Tools Track, and Education and Training Track. Co-located with REFSQ 2024. Winterthur, Switzerland, April 8, 2024.}

\title{Which AI Technique Is Better to Classify Requirements? An Experiment with SVM, LSTM, and ChatGPT}

\author[1]{Abdelkarim El-Hajjami}[
orcid=0009-0004-7053-3264,
email=abdelkarim.el-hajjami@univ-paris1.fr,
]
\address[1]{Paris 1 Panthéon–Sorbonne University, Paris, France}
\author[1]{Nicolas Fafin}[
email=nicolas.fafin@etu.univ-paris1.fr,
]
\author[1]{Camille Salinesi}[%
orcid=0000-0002-1957-0519,
email=camille.salinesi@univ-paris1.fr,
]
\cortext[1]{Corresponding author.}
\begin{abstract}
Recently, Large Language Models like ChatGPT have demonstrated remarkable proficiency in various Natural Language Processing tasks. Their application in Requirements Engineering, especially in requirements classification, has gained increasing interest. This paper reports an extensive empirical evaluation of two ChatGPT models, specifically gpt-3.5-turbo, and gpt-4 in both zero-shot and few-shot settings for requirements classification. The question arises as to how these models compare to traditional classification methods, specifically Support Vector Machine and Long Short-Term Memory. Based on five different datasets, our results show that there is no single best technique for all types of requirement classes. Interestingly, the few-shot setting has been found to be beneficial primarily in scenarios where zero-shot results are significantly low.
\end{abstract}

\begin{keywords}
Large Language Models \sep ChatGPT \sep Requirements Classification
\end{keywords}

\maketitle

\section{Introduction}
Artificial Intelligence (AI) presents an exciting opportunity for Requirements Engineering (RE). As many publications have already shown, AI is enhancing the precision and efficiency of many RE tasks from elicitation to negotiation, and formalization. This paper focuses on requirements classification, particularly the challenge of differentiating between Functional (FR) and Non-Functional Requirements (NFR), a significant matter in both academic research and practical industry applications.

The distinction between FR, and NFR is subject to ongoing debate within the community, and there is no consensus on their definitions \cite{chungetal2012} \cite{eckhardtetal2016} \cite{lietal2014}. We believe that employing Large Language Models (LLMs) as requirements classifiers will shed further light on this scientific discussion.

Software development projects often involve managing a multitude of requirements, making it essential for software professionals to effectively organize and prioritize them. One key aspect of this process is the classification of requirements into FR and NFR. However, manually categorizing each requirement is a labor-intensive process. Therefore, there is a compelling need for an automated method that can accurately and efficiently identify FR and NFR.

Our research is grounded upon Dalpiaz et al. version. \cite{dalpiazetal2019} of Kurtanović and Maalej’s requirements classification approach \cite{kurtanovicmaalej2017}. Since the original classifier by Kurtanović and Maalej is not available online, we rely on Dalpiaz et al.'s reconstruction, emphasizing the differentiation of FR and NFR using high-dimensional (500) word level features.

Building upon Dalpiaz et al.'s work with SVM, our research explores the Long Short-Term Memory (LSTM) model and two ChatGPT versions, gpt-3.5-turbo and gpt-4 by OpenAI, to align with our specific research objectives. Based on this, our research aims to address the following research questions:
\begin{itemize}
    \item \textbf{RQ1:} What is the best technique for requirements classification between SVM, LSTM and ChatGPT?
    \item \textbf{RQ2:} What are the performance differences between GPT-4 and 3.5?
    \item \textbf{RQ3:} What are the performance differences between Zero-Shot and Few-Shot settings?
\end{itemize}
The rest of the paper is organized as follows: Section II outlines the methodology, providing insights into our research approach. Section III presents the analysis of our experimental results. Section IV comprises the threats to validity of our research, while Section VI presents the conclusion, synthesizing our findings and suggesting future directions in the field.
\section{Methodology}
\subsection{The classification problem}
In our methodology, we follow the modeling paradigm of Li et al. \cite{lietal2014} as further refined by Dalpiaz et al. The approach categorizes requirements considering two aspects in any requirements: the functional aspect and the quality aspect. A requirement has a functional aspect when it specifies a functional goal or a functional constraint. In parallel, the quality aspects of a requirement include quality goals and quality constraints.

Acknowledging the possibility that a requirement may encompass both functional and quality aspects, Dalpiaz et al. subsequently formulated four distinct binary classification problems:
\begin{itemize}
  \item \textbf{IsFunctional:} does a requirement possess functional aspects?
  \item \textbf{IsQuality:} does a requirement possess quality aspects?
  \item \textbf{OnlyFunctional:} does a requirement possess only functional aspects?
  \item \textbf{OnlyQuality:} does a requirement possess only quality aspects?
\end{itemize}

\subsection{The Datasets}
While our initial goal was to utilize the exact same datasets as the ones curated by Dalpiaz et al., we had access to only five public requirements datasets out of the eight used by Dalpiaz et al. in their experiments: PROMISE, Dronology, ReqView, Leeds Library and WASP \cite{dalpiazetal2019materials}.

Table \ref{tab:datasets} indicates the number of requirements and their distribution among the four classes.
\begin{table}[h]
\centering
\footnotesize
\caption{Datasets overview}
\begin{tabular}{lccccc}
\toprule
\textbf{Dataset} & \textbf{Rows} & \textbf{IsFunctional} & \textbf{IsQuality} & \textbf{OnlyFunctional} & \textbf{OnlyQuality} \\
\midrule
PROMISE & 625 & 310 & 382 & 230 & 302 \\
Dronology & 97 & 94 & 28 & 68 & 2 \\
ReqView & 87 & 75 & 32 & 54 & 11 \\
Leeds Library & 85 & 44 & 61 & 23 & 40 \\
WASP & 62 & 55 & 19 & 42 & 6 \\
\midrule
\textbf{Totals} & 956 & 578 & 522 & 417 & 361 \\
\bottomrule
\end{tabular}
\label{tab:datasets}
\end{table}

Table \ref{tab:datasets} reveals a clear imbalance in the datasets across the different classes. The "IsFunctional" class has 578 instances compared to "OnlyQuality" with only 361 instances. This pattern of imbalance extends across all datasets, with "IsFunctional" and "IsQuality" classes being more common than "OnlyFunctional" and "OnlyQuality" classes.
\newpage
\subsection{The Evaluated Models}
\subsubsection{SVM}
We directly used the datasets from the work of Dalpiaz et al. for their SVM-based classification approach \cite{dalpiazetal2019materials}, which employs 500 word-level features such as text n-grams or Part-of-Speech (POS) n-grams. We trained the model on 75\% of the PROMISE dataset and evaluated its performance on the global evaluation dataset, which was formed by combining the individual test sets—specifically, the remaining 25\% of the PROMISE dataset, along with the datasets from Dronology, ReqView, Leeds Library, and WASP. This configuration allowed us to assess the model's performance on diverse and unseen data.

\subsubsection{LSTM}
LSTM, a type of Recurrent Neural Network (RNN), is especially effective for handling sequential data, making it relevant for requirements classification problems. Before evaluating the capabilities of ChatGPT in requirements classification, it was imperative to benchmark its performance against a well-established model known for its sequential data processing capabilities.
The LSTM model architecture in this study includes an embedding layer that converts words into numerical vectors, a spatial dropout layer for regularization to prevent overfitting, an LSTM layer that processes text sequences and captures word dependencies, and a dense layer with a sigmoid activation for outputting a probability score indicating the likelihood of a requirement being in the positive class.

Mirroring the approach adopted for the SVM model, we trained the LSTM on 75\% of the PROMISE dataset and assessed its performance on the global evaluation dataset.

\subsubsection{ChatGPT}
ChatGPT, developed by OpenAI, is a Generative Pre-trained Transformer model designed to generate human-like text based on the prompts it receives.

In our study, we used:
\begin{enumerate}
  \item \textbf{gpt-3.5-turbo:} As of the date of our experimentation, October 20, 2023, this is the latest and most advanced model in the GPT-3.5 series. Its training data extends up to September 2021.
  \item \textbf{gpt-4:} OpenAI's newest model, trained until September 2021.
\end{enumerate}

\paragraph{Prompt Engineering}
Prompt engineering is a critical aspect of interacting with ChatGPT models. Crafting effective prompts can significantly influence the model's responses, particularly when precise or specific outputs are desired.

In the context of our study, two major strategies were employed: zero-shot prompting and few-shot prompting \cite{openai2020}. Both of these strategies aimed to guide the model in correctly classifying a requirement, but they approach the task differently.

\paragraph{Zero-Shot Prompting}
In Zero-shot prompting, models are given the task without the benefit of specific contextual examples in the prompt. As such, the model draws purely from its vast pre-trained knowledge and the immediate context provided in the prompt. Zero-shot scenarios offer a unique insight into the model's innate understanding of the task and its ability to generalize from its training data without needing explicit examples to guide its response.

For the zero-shot setting in our study, our prompts were engineered following these three basic principles:
\begin{enumerate}
  \item  Each prompt should include all necessary details to get more relevant answers;
  \item Each prompt should ask the model to adopt a software requirements expert persona.
  \item The words used in each prompt should come from the Dalpiaz et al. reference paper.
\end{enumerate}

\newpage
For example, the designed zero-shot prompt for the "IsFunctional" classification is:
\begin{mdframed}[
    backgroundcolor=black!10,
    linewidth=0.5pt,
    linecolor=black,
    topline=true,
    bottomline=true,
    leftline=true,
    rightline=true,
    innerleftmargin=8pt,
    innerrightmargin=8pt,
    innertopmargin=8pt,
    innerbottommargin=8pt
]
    \footnotesize
    \setlength{\baselineskip}{1.4\baselineskip}
    
    You are a software requirements expert tasked with categorizing software requirements into:\newline
    IsFunctional: if the requirement possesses functional aspects (functional goals or functional constraints).\newline
    IsQuality: if the requirement possesses quality aspects (quality goals or quality constraints).\newline
    OnlyFunctional: if the requirement possesses only functional aspects, with no quality aspects.\newline
    OnlyQuality: if the requirement possesses only quality aspects, with no functional aspects.\newline
    Given the requirement below, determine if it falls under the class "IsFunctional"\newline
    Requirement: "[requirement\_text]"\newline
    Respond as "yes" or "no".
\end{mdframed}

\paragraph{Few-Shot Prompting}
Few-shot prompting is another approach to guide the model towards desired outputs by providing a limited number of example inputs and their corresponding outputs. By providing these examples, the intention is to give the model a clear context of what is expected from it, thus aiding it in generating more accurate and task-specific responses.

In our study, few-shot prompting was applied with adherence to a set of structured guidelines. Firstly, we adopted the same principles as those used in the zero-shot setting. Additionally, for the selection of examples, we ensured that all were sourced from the PROMISE training set. A balanced representation was a key focus; for each class — "IsFunctional", "IsQuality", "OnlyFunctional", and "OnlyQuality" — we curated two examples each of the positive and negative cases. This approach was designed to ensure that the model was exposed to a diverse range of scenarios, thereby providing a comprehensive understanding of the classification nuances. For instance, in the case of "IsFunctional", the positive instances included one scenario that combined both "IsFunctional" and "IsQuality", and another that incorporated "IsFunctional" without "IsQuality".

It is important to note that ChatGPT was also evaluated on the same global evaluation dataset as used for SVM and LSTM, which includes the 25\% of the PROMISE dataset and the entirety of the datasets from Dronology, ReqView, Leeds Library, and WASP. The examples for few-shot prompting were specifically chosen from the PROMISE training set to maintain consistency.

\paragraph{Querying ChatGPT}
The OpenAI API was used to query ChatGPT. Every individual instance of each test set was provided to ChatGPT in one API call. The parameters of ChatGPT API calls were set as follows:

\begin{itemize}
  \item \textbf{Temperature:} We set the temperature parameter to 0 in order to minimize randomness in outputs and enhance the reproducibility of our evaluation results. However, due to the non-deterministic nature of ChatGPT, some variability may still occur.
  \item \textbf{Other parameters:} Beyond the temperature setting, all other parameters were set at their default values.
\end{itemize}

While running the experiments, we encountered multiple server-side
errors (e.g., BadGateway – HTTP 502, ServiceUnavailableError – HTTP 500). We incorporated exception handling in our approach in order to handle them.

To provide a clear understanding of the implementation details and ensure transparency and reproducibility, the code used in this study has been made publicly available \cite{materials}.
\section{Experimental Results Analysis}
\subsection{The Evaluation Metric}
In the context of our requirements classification task, which is characterized by a pronounced imbalance in the dataset as shown in Table \ref{tab:datasets}, the selection of an appropriate evaluation metric is important. Dalpiaz et al. have previously employed a range of metrics including precision, recall, F1-score, and AUC to assess model performance. Since ChatGPT does not offer probability distribution or score confidence which are essential for the calculation of the AUC metric, we cannot employ this metric in our evaluation.
The F1-score, while commonly used, assumes that precision and recall are of equal importance by attributing them equal weight. This assumption does not hold in our context, especially considering that achieving high recall is more challenging and crucial in manual processes. Typically, it is more feasible to manually reject a false positive identified by a tool than to manually find a true positive within the tool's input.
Consequently, we propose the use of the ${F_\beta}$ score, which allows us to weight recall over precision to reflect its greater importance. The value of $\beta$ should be empirically determined based on the inverse frequency of the target class within the dataset \cite{berry2021}. As explained by Berry \cite{berry2021}, the inverse frequency indicates the average number of items that must be examined in the search space to find one true positive, providing a lower bound estimate for ${F_\beta}$.

Table \ref{tab:beta_values} is constructed based on the inverse frequencies derived from the datasets, illustrating the ${\beta}$ values for each class.
\begin{table}[h]
\centering
\footnotesize
\caption{Inverse Frequency of Classes as \(\beta\) Values}
\label{tab:beta_values}
\begin{tabular}{lcccc}
\toprule
\textbf{Test set}        & \textbf{\(\beta_{\text{IsFunctional}}\)} & \textbf{\(\beta_{\text{IsQuality}}\)} & \textbf{\(\beta_{\text{OnlyFunctional}}\)} & \textbf{\(\beta_{\text{OnlyQuality}}\)} \\ \hline
PROMISE Test       & 2.09                          & 1.53                       & 2.87                            & 1.91                         \\
Dronology      & 1.03                          & 3.46                       & 1.43                            & 48.50                        \\
ReqView        & 1.16                          & 2.72                       & 1.61                            & 7.91                         \\
Leeds Library  & 1.93                          & 1.39                       & 3.70                            & 2.13                         \\
WASP           & 1.13                          & 3.26                       & 1.48                            & 10.33                        \\
\hline
Global         & 1.43                          & 2.02                       & 2.02                            & 3.49                        \\
\bottomrule
\end{tabular}
\end{table}
\subsection{RQ1: What is the best technique for requirements classification between SVM, LSTM and ChatGPT?}
Table \ref{tab:rq1} provides a comprehensive evaluation of the global ${F_\beta}$ scores for SVM, LSTM, GPT-3.5, and GPT-4 models, including both Zero-Shot and Few-Shot configurations, across four binary requirements classifications: "IsFunctional", "IsQuality", "OnlyFunctional", and "OnlyQuality".
\begin{table}[h!]
\centering
\footnotesize
\caption{The global ${F_\beta}$ score of SVM, LSTM, and GPT configurations}
\begin{tabular}{l|c|c|c|c|c|c}
\toprule
\textbf{Classification} & \textbf{SVM} & \textbf{LSTM} & \multicolumn{2}{c|}{\textbf{GPT-3.5}} & \multicolumn{2}{c}{\textbf{GPT-4}} \\
\cline{4-7}
& & & \textbf{Zero-Shot} & \textbf{Few-Shot} & \textbf{Zero-Shot} & \textbf{Few-Shot} \\
\midrule
IsFunctional & 0.809 & 0.774 & 0.848 & \textbf{0.899} & 0.895 & 0.888 \\
IsQuality & 0.691 & \textbf{0.708} & 0.435 & 0.687 & 0.610 & 0.493 \\
OnlyFunctional & 0.533 & 0.588 & 0.000 & 0.692 & \textbf{0.872} & 0.843 \\
OnlyQuality & 0.686 & 0.643 & 0.000 & \textbf{0.732} & 0.554 & 0.497 \\
\bottomrule
\end{tabular}
\label{tab:rq1}
\end{table}

The ${F_\beta}$ scores in Table \ref{tab:rq1} reveal that no single technique consistently outperforms the others across all four classifications. Rather, the optimal classification technique is contingent upon the specific requirements class pursued. For instance, in the "IsFunctional" classification, GPT-3.5 and GPT-4 are highly competitive in both Zero-Shot and Few-Shot settings, with GPT-3.5 Few-Shot reaching an ${F_\beta}$ score of 0.899. Meanwhile, LSTM leads in the "IsQuality" classification with a score of 0.708. The "OnlyFunctional" classification is best addressed by the GPT-4 Zero-Shot model, which scores 0.872. Lastly, for requirements that are solely quality-centric ("OnlyQuality"), the GPT-3.5 Few-Shot setting also shows a commendable ${F_\beta}$ score of 0.732.

An additional nuance to consider regarding the performance of the LSTM model is that although it is the best performer in the "IsQuality" classification, its effectiveness is not consistent across all the individual test sets. Indeed, its performance is particularly good on the PROMISE test set, likely due to being trained on similar data. However, its effectiveness drops when applied to other test sets, as illustrated in Table $\ref{tab:lstm_evaluation_isq}$, indicating a limitation in its ability to generalize.
\begin{table}[h!]
\centering
\footnotesize
\caption{LSTM evaluation metrics for "IsQuality" classification}
\begin{tabular}{lccc}
\toprule
\textbf{Dataset} & \textbf{Precision} & \textbf{Recall} & \textbf{F$_\beta$} \\
\midrule
PROMISE Test & 0.913 & 0.931 & 0.925 \\
Dronology & 0.277 & 0.643 & 0.584 \\
ReqView & 0.375 & 0.750 & 0.670 \\
Leeds Library & 0.623 & 0.541 & 0.566 \\
WASP & 0.311 & 0.737 & 0.659 \\
\bottomrule
\end{tabular}
\label{tab:lstm_evaluation_isq}
\end{table}

A further observation from Table \ref{tab:onlyquality-fbeta-comparison} is that none of the models were able to effectively classify "OnlyQuality" requirements in the Dronology test set except for the LSTM model. This issue likely results from the dataset's significant imbalance, with only two "OnlyQuality" samples present. The absence of a substantial number of such samples means that failing to identify these few instances translates directly to a score of zero. While the ${F_\beta}$ score is designed to provide a more nuanced evaluation in the presence of class imbalance by emphasizing recall, this example shows that even a well-chosen metric may not be sufficient in the context of highly imbalanced datasets.
\begin{table}[h!]
\centering
\footnotesize
\caption{The "OnlyQuality" ${F_\beta}$ performance metrics comparison}
\begin{tabular}{l|c|c|c|c|c|c}
\toprule
\textbf{Test set} & \textbf{SVM} & \textbf{LSTM} & \multicolumn{2}{c|}{\textbf{GPT-3.5}} & \multicolumn{2}{c}{\textbf{GPT-4}} \\
\cline{4-7}
& & & \textbf{Zero-Shot} & \textbf{Few-Shot} & \textbf{Zero-Shot} & \textbf{Few-Shot} \\
\midrule
PROMISE Test & 0.861 & 0.887 & 0.000 & 0.727 & 0.576 & 0.593 \\
\textbf{Dronology} & \textbf{0.000} & 0.989 & \textbf{0.000} & \textbf{0.000} & \textbf{0.000} & \textbf{0.000} \\
ReqView & 0.723 & 0.347 & 0.000 & 0.805 & 0.638 & 0.368 \\
Leeds Library & 0.460 & 0.384 & 0.000 & 0.718 & 0.623 & 0.500 \\
WASP & 0.331 & 0.322 & 0.000 & 0.498 & 0.502 & 0.168 \\
\bottomrule
\end{tabular}
\label{tab:onlyquality-fbeta-comparison}
\end{table}

In summarizing our findings, it is clear that there is no single best technique for all requirements classifications. The best technique varies depending on the specific class of requirements that you need to identify.
\subsection{RQ2: What are the performance differences between GPT-4 and 3.5?}
\begin{table}[h!]
\centering
\footnotesize
\caption{The global ${F_\beta}$ score of GPT configurations}
\begin{tabular}{l|c|c|c|c}
\toprule
\textbf{Classification} & \multicolumn{2}{c|}{\textbf{GPT-3.5}} & \multicolumn{2}{c}{\textbf{GPT-4}} \\
\cline{2-5}
& \textbf{Zero-Shot} & \textbf{Few-Shot} & \textbf{Zero-Shot} & \textbf{Few-Shot} \\
\midrule
IsFunctional & 0.848 & \textbf{0.899} & 0.895 & 0.888 \\
IsQuality & 0.435 & \textbf{0.687} & 0.610 & 0.493 \\
OnlyFunctional & 0.000 & 0.692 & \textbf{0.872} & 0.843 \\
OnlyQuality & 0.000 & \textbf{0.732} & 0.554 & 0.497 \\
\bottomrule
\end{tabular}
\label{tab:gpt-comparison}
\end{table}
In an analytical comparison of GPT-3.5 and GPT-4 based on the given performance metrics as presented in Table \ref{tab:gpt-comparison}, one can observe that GPT-3.5 outperforms GPT-4 in three out of four scenarios. This is particularly notable in the "IsQuality" and "OnlyQuality" classifications, where GPT-3.5 has higher ${F_\beta}$ scores. GPT-4 only stands out is in the "OnlyFunctional" classification, demonstrating a significant improvement over GPT-3.5 in both zero-shot and few-shot settings. Therefore, if the classification of purely functional requirements is critical and the budget allows for the higher cost of GPT-4, it would be the recommended choice.
\subsection{RQ3: What are the performance differences between Zero-Shot and Few-Shot settings?}
Table \ref{tab:delta_comparison} presents the score deltas, computed as the few-shot global ${F_\beta}$ score minus the zero-shot global ${F_\beta}$ score, for GPT-3.5 and GPT-4 across our four classifications. This metric serves as an indicator of the added value from few-shot learning over an initial zero-shot baseline.
\begin{table}[h!]
\centering
\footnotesize
\caption{The ${F_\beta}$ score delta between Few-Shot and Zero-Shot for ChatGPT}
\begin{tabular}{l|ccc|ccc}
\toprule
\textbf{Classification} & \multicolumn{3}{c|}{\textbf{GPT-3.5}} & \multicolumn{3}{c}{\textbf{GPT-4}} \\
\cmidrule(lr){2-4} \cmidrule(lr){5-7}
& Zero-Shot & Few-Shot & \textbf{Delta} & Zero-Shot & Few-Shot & \textbf{Delta} \\
\midrule
IsFunctional & 0.848 & 0.899 & 0.051 & 0.895 & 0.888 & -0.007 \\
IsQuality & 0.435 & 0.687 & 0.252 & 0.610 & 0.493 & -0.117 \\
OnlyFunctional & \textbf{0.000} & \textbf{0.692} & \textbf{0.692} & 0.872 & 0.843 & -0.029 \\
OnlyQuality & \textbf{0.000} & \textbf{0.732} & \textbf{0.732} & 0.554 & 0.497 & -0.057 \\
\bottomrule
\end{tabular}
\label{tab:delta_comparison}
\end{table}

The deltas show that few-shot setting does not consistently improve upon the zero-shot baseline, with marginal declines in performance across most classifications. Conversely, GPT-3.5 displays significant positive deltas in "OnlyFunctional" and "OnlyQuality" classifications, indicating that few-shot learning has a pronounced beneficial impact when the zero-shot performance is weak. These insights suggest that few-shot learning can be particularly valuable in scenarios where a model struggles to perform adequately without prior examples, as it can lead to marked performance gains and help models like GPT-3.5 overcome initial deficiencies in understanding and classifying requirements.
\section{Threats to Validity}
In our study, we identified several threats to validity across different dimensions. Internally, we noted biases such as the sensitivity of ChatGPT's responses to prompt variations and the non-deterministic nature of its output, which could vary slightly even with identical prompts. The choice and representation of few-shot examples and the potential impact of server errors also posed challenges to the reliability of our findings. Externally, the use of specific datasets and the subjectivity of tagger annotations could introduce biases. Regarding construct validity, we acknowledged that traditional metrics might not fully reflect the model's capabilities in classifying requirements, accentuating the importance of interpretability. Finally, our conclusion validity faced threats from unbalanced datasets, which could bias our results.
\section{Related Work}
The work of Rashwan et al. \cite{rashwanetal2013} introduced a novel approach by developing a new corpus with annotations for different types of NFR based on a requirements ontology and employing a SVM classifier to categorize requirements sentences into different ontology classes automatically. The proposed approach showed promising results in two different software requirements specification corpora, underlining the importance of semantic analysis and ontological representation.

Ray et al. \cite{rayetal2023} introduced the aeroBERT-Classifier, a specialized model designed for the aerospace domain. The model's architecture, leveraging a domain-adapted BERT, underscores the merits of industry-specific adaptations. Their comparative analysis against models like GPT-2 and Bi-LSTM further delineates the robustness of transformer architectures in similar tasks.

In a distinct contribution, Rahimi et al. \cite{rahimietal2020} proposed an ensemble approach combining a suite of machine learning classifiers, including Naïve Bayes, SVM, Decision Tree, Logistic Regression, and SVC. Their ensemble method achieved a remarkable 99.45\% accuracy in classifying Functional Requirements, affirming the efficacy of harnessing multiple models to optimize classification outcomes.

Lastly, adding a linguistic dimension, Yucalar \cite{yucalar2023} introduced BERTurk, a model specifically fine-tuned for classifying software requirements in the Turkish language. Through rigorous empirical validation, Yucalar established that BERTurk achieves a commendable 95\% F1-score in differentiating between functional and non-functional requirements. This research accentuates the value of integrating linguistic nuances with state-of-the-art NLP techniques within the sphere of RE.
\section{Conclusion and Future Work}
In this study, we conducted a comprehensive assessment of two ChatGPT models: gpt-3.5-turbo, and gpt-4. We evaluated these models in both zero-shot and few-shot settings, comparing them against established methods such as SVM and LSTM in requirements classification. Our findings show that there is no single best technique for all requirements classifications. The best technique varies depending on the specific requirement classification. For instance, 
GPT-3.5 Few-Shot configuration leads in the "IsFunctional" and "OnlyQuality" classifications, LSTM model performs best in the "IsQuality" classification, and GPT-4 Zero-Shot setting stands out in the "OnlyFunctional" classification. Our results also indicate that GPT-3.5 is generally more effective than GPT-4, except when it comes to "OnlyFunctional" requirements classification, where GPT-4's higher cost may be justified by its enhanced performance.
Interestingly, the few-shot setting has been found to be beneficial primarily in scenarios where zero-shot performance is notably weak.
From a practical perspective, these findings suggest that for "IsFunctional" and "OnlyQuality" classifications, the default choice should be GPT-3.5 Few-Shot.
In scenarios where the classification of "OnlyFunctional" requirements is paramount and resources are available, GPT-4 emerges as a reasonably good cold start. If budget constraints are a factor, GPT-3.5 Few-Shot remains a viable alternative. For "IsQuality" classification, LSTM stands out as the most effective tool, however it needs to be trained.

In the future, we intend to undertake  a series of rigorous comparisons of ChatGPT with other LLMs, such as Llama-2 and Mistral, in performing RE tasks. Such analyses could offer valuable insights into the relative advantages and limitations of each model for various RE tasks. To complement this research direction, it is also essential to build high-quality benchmark requirements datasets for training and comprehensively evaluating LLMs on a wider range of RE tasks.
%
%
%
%
\bibliography{references}

\end{document}